\documentclass{article}

\usepackage{microtype}
\usepackage{graphicx}
\usepackage{soul}
\usepackage{subcaption}
\usepackage{booktabs} 
\usepackage{multirow}
\usepackage{enumitem}
\usepackage{array}

\usepackage{hyperref}
\usepackage[table]{xcolor}

\definecolor{mygray}{gray}{0.85}
\definecolor{mypink}{rgb}{1, 0.8, 0.8}

\usepackage[preprint]{icml2026}

\usepackage{amsmath}
\usepackage{amssymb}
\usepackage{mathtools}
\usepackage{amsthm}

\usepackage{algorithm}
\usepackage{algorithmic}
\usepackage{amsmath}

\usepackage{svg}

\usepackage[capitalize,noabbrev]{cleveref}

\theoremstyle{plain}

\theoremstyle{definition}

\theoremstyle{remark}

\usepackage[textsize=tiny]{todonotes}

\newcommand{\RETURN}{\STATE \textbf{return}}
\newcommand{\masked}{\odot}
\newcommand{\eg}{{\it e.g.}, }
\newcommand{\circleNum}[1]{\tikz[baseline=(char.base)]\node[circle, fill=black, text=white, inner sep=1pt, minimum size=0.6em, font=\scriptsize] (char) {#1};}

\icmltitlerunning{CHESS: Context-aware Hierarchical Efficient Semantic Selection for Long-Context LLM Inference}

\usepackage{microtype}

\begin{document}

\twocolumn[
  \icmltitle{CHESS: Context-aware Hierarchical Efficient Semantic Selection for Long-Context LLM Inference}


  \icmlsetsymbol{equal}{*}

  \begin{icmlauthorlist}
    \icmlauthor{Chao Fei}{kaust}
    \icmlauthor{Guozhong Li}{kaust}
    \icmlauthor{Chenxi Liu}{cair}
    \icmlauthor{Panos Kalnis}{kaust}
  \end{icmlauthorlist}

  \icmlaffiliation{kaust}{King Abdullah University of Science and Technology (KAUST), Thuwal, Saudi Arabia}
  \icmlaffiliation{cair}{Centre for Artificial Intelligence and Robotics, Hong Kong Institute of Science \& Innovation, Chinese Academy of Sciences, Hong Kong, China}

  \icmlcorrespondingauthor{Panos Kalnis}{panos.kalnis@kaust.edu.sa}

  \icmlkeywords{Large Language Models, KV Cache, Long Context, Attention Mechanism, Uncertainty}

  \vskip 0.3in
]



\printAffiliationsAndNotice{}  

\begin{abstract}
    Long-context LLMs demand accurate inference at low latency, yet decoding becomes primarily constrained by KV cache as context grows.
    Prior pruning methods are largely context-agnostic: their token selection ignores step-wise relevance and local semantics, which undermines quality.
    Moreover, their irregular accesses and selection overheads yield only limited wall-clock speedups.
    To address this, we propose \textbf{CHESS}, an \textit{algorithm-system co-design} KV-cache management system.
    Algorithmically, CHESS introduces a context-aware, hierarchical selection policy that dynamically reconstructs a coherent context for the current decoding.
    System-wise, coarse granularity selection eliminates expensive data movement, fully realizing practical acceleration from theoretical sparsity.
    Extensive evaluations demonstrate that CHESS surpasses Full-KV quality using only \textbf{1\%} of the KV cache, delivers low-latency stable inference with up to \textbf{4.56$\times$} higher throughput, and consistently outperforms other strong baselines.
    Code is available at \href{https://anonymous.4open.science/r/CHESS-9958/}{https://anonymous.4open.science/r/CHESS/}.
\end{abstract}

\section{Introduction}
\label{sec:intro}

\begin{figure}[!htb]
    \centering
    \includegraphics[width=\linewidth]{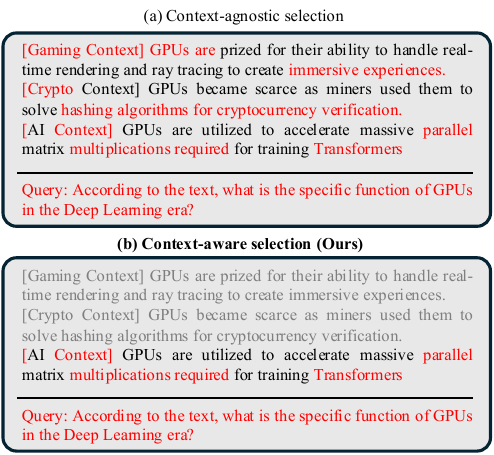}
    \caption{\textbf{Context-agnostic vs. context-aware KV selection.} Red indicates critical tokens, while grey denotes ignored tokens. 
    (a) \textbf{Context-agnostic (\eg SnapKV)}: Preserve the most important tokens based on attention scores. 
    (b) \textbf{Context-aware (Ours)}: Adaptively select segments that are semantically relevant to the current generation and retain local context.}
    \label{fig:case-study}
\end{figure}

Large language models (LLMs) now power a broad range of applications, from short-form dialogue and QA~\cite{DBLP:conf/emnlp/HwangKBLBJ23} to agent-based workflows~\cite{DBLP:conf/iclr/ZhangXYTCCZCHWZ25}, data-processing pipelines~\cite{DBLP:journals/pvldb/NarayanCOR22}, and long-form generation~\cite{DBLP:conf/acl/KimK25}, and have become integral to daily use~\cite{DBLP:conf/acl/0002WLC025}. 
These scenarios continually expand context windows, making KV-cache management the primary bottleneck for inference.
During decoding, each new token attends to an expanding prefix, requiring $O(L)$ KV reads per layer from off-chip memory into on-chip SRAM.
This I/O makes inference memory-bandwidth-bound and causes latency to scale linearly with context length.
Consequently, the fundamental challenge lies in overcoming this memory-bandwidth bottleneck to achieve efficient long-context inference while preserving the generation quality.

A promising direction is to retain only a critical subset of tokens in the KV cache. 
Prior work~\cite{zhang2023h2oheavyhitteroracleefficient,ge2024modeltellsdiscardadaptive} shows that a small fraction of tokens contributes most to generation, and enables substantial memory and latency reductions with little quality loss. 
However, existing methods typically adhere to a {context-agnostic} paradigm. 
As illustrated in Figure~\ref{fig:case-study}(a), these approaches often retain globally frequent tokens regardless of their relevance to the current query, failing to adapt to dynamic semantic contexts. 
Furthermore, the implementation of system-algorithm co-design to fully realize hardware efficiency remains a largely underexplored direction. 
Specifically, regarding system implementation, substantial challenges persist: dynamic cache updates may trigger expensive memory copy operations, while irregular selection logic can significantly hinder the parallelization required for large batch sizes.

In this work, we propose \textbf{CHESS} (\textbf{C}ontext-aware \textbf{H}ierarchical \textbf{E}fficient \textbf{S}emantic \textbf{S}election), a novel KV-cache management system for long-context LLM inference. 
Our design is driven by the key observation that token importance is inherently \textbf{context-dependent} and shifts during decoding, as illustrated in Figure~\ref{fig:case-study}(b). 
Accordingly, CHESS adaptively reconstructs a semantically relevant context for the current decoding step to minimize input processing overhead while preserving precision. 
\circleNum{1} Algorithmically, CHESS first utilizes {hierarchical semantic selection} to efficiently identify relevant context blocks. 
CHESS further incorporates an {uncertainty-aware backtracking} mechanism that monitors generation quality, dynamically retrieving previously pruned information to ensure robustness.
\circleNum{2} System-wise, to fully realize the efficiency of these algorithmic designs, we implement a {zero-copy inference engine} atop PagedAttention~\cite{kwon2023efficientmemorymanagementlarge}. 
Instead of performing expensive physical data movement, CHESS selectively manages memory by manipulating logical page indices. 
CHESS minimizes system overhead by leveraging high-performance GEMM optimizations for the selection algorithm and encapsulating the entire workflow into CUDA Graphs.

We evaluate CHESS on LongBenchV2~\cite{bai2025longbenchv2deeperunderstanding} for quality and on synthetic data of varying lengths for efficiency.
Under merely $1\%$ of the KV cache, CHESS outperforms the full-context baseline, as our dynamic selection effectively filters out irrelevant noise. 
These algorithmic gains translate directly into system performance: CHESS exhibits robust scalability in long-context generation, consistently surpassing state-of-the-art methods and achieving up to {4.56$\times$} higher throughput than full KV in large-batch scenarios.

In all, our contributions are summarized as follows:
\begin{itemize}[nosep, leftmargin=*]
    \item We observe the dynamic nature of token importance during decoding and propose a {context-aware} selection paradigm to adaptively construct relevant context that supports the current generation, while discarding irrelevant tokens that serve as noise.
    \item We propose {CHESS}, an algorithm-system co-design system. It not only maintains high algorithmic accuracy but also implements specialized kernels to translate sparsity into practical system efficiency.
    \item Extensive experiments demonstrate that CHESS maintains generation quality on LongBenchV2 with just 1\% of the KV cache, while achieving up to $4.56\times$ throughput improvement on synthetic data, consistently exceeding other baselines.
\end{itemize}

\section{Challenges of Long-Context Decoding}
\label{sec:background}

\begin{figure}[t]
    \centering
    \includegraphics[width=\linewidth]{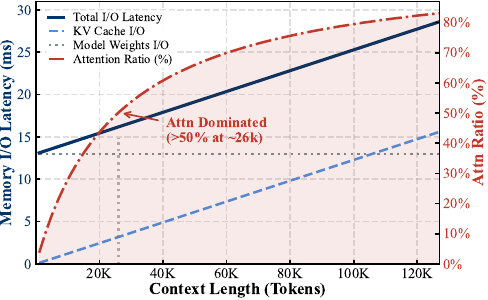}
    \caption{Escalating latency and dominant attention computation in long-context decoding.
    }
    \label{fig:latency_computation}
\end{figure}

\begin{figure*}[!htbp]
    \centering
    \includegraphics[width=\textwidth]{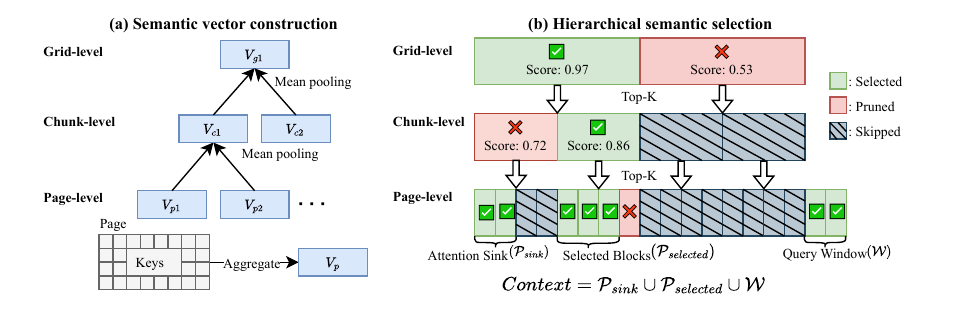}
    \vspace{-1cm}
    \caption{Overview of the CHESS System Architecture. The system maintains a hierarchical view (Grid, Chunk, Page) over the physical KV cache to enable context-aware selection. The final context is reconstructed by combining semantically selected pages with attention sinks and the local query window.}
    \label{fig:system_arch}
\end{figure*}

\subsection{Scaling Constraints}
\paragraph{Memory wall.} 
Long contexts make decoding {KV-cache–bound}.
The KV footprint grows roughly linearly with sequence length: \eg with Qwen3-8B (bfloat16)~\cite{yang2025qwen3technicalreport}, a single 12K-token sequence occupies about {1.5\,GB} of KV memory.
By contrast, on-chip SRAM is scarce (A100 L2 cache: {40\,MB}), so each decoding step must repeatedly read \(O(L)\) keys/values per layer from HBM to SRAM.
This off-chip traffic saturates memory bandwidth and stalls the pipeline.

\paragraph{Compute wall.}
As context length \(L\) grows, attention operates over an expanding prefix and the per-step FLOPs increase roughly linearly with \(L\).
Empirically (Figure~\ref{fig:latency_computation}), {KV-cache I/O} (blue dashed) rises approximately linearly with \(L\) while {model-weights I/O} (gray dotted) stays nearly flat, so total memory I/O (blue solid) increases with length.
Meanwhile, the {attention ratio} (red) climbs with \(L\) and exceeds \(50\%\) at around \(26\mathrm{K}\) tokens, trending toward dominance at longer contexts.
Together, these trends indicate a dual pressure in long-context decoding: memory bandwidth is driven by KV I/O growth, and attention compute increasingly dominates end-to-end latency.

\subsection{Gaps in Prior Approaches}
While various long-context methods theoretically reduce memory footprint, their integration into high-performance serving systems exposes structural inefficiencies. 
We identify three bottlenecks and the design opportunities.

\paragraph{Metric overhead breaks kernel fusion.}
Many sparsification methods~\cite{zhang2023h2oheavyhitteroracleefficient,rocketKV,li2024snapkvllmknowslooking} rely on exact attention scores to identify important tokens. 
However, this approach is not naturally aligned with modern optimized kernels, which achieve acceleration specifically by avoiding the storage of intermediate attention maps. 
Consequently, accessing these scores requires additional memory operations, introducing I/O overhead that partially diminishes the efficiency gains brought by sparsity.

\paragraph{Token granularity inefficiency.}
Token-level selection methods~\cite{xiao2024efficientstreaminglanguagemodels, keydiff} often necessitate physical data movement to maintain memory continuity, consuming additional bandwidth. 
Crucially, these fine-grained approaches present a structural mismatch with modern block-based memory managers. 
Operating at the individual token level disrupts the block abstraction, which can limit the effectiveness of optimized block-based kernels and increase management complexity.

\paragraph{Context-agnostic selection.}
Beyond system costs, many approaches perform context-agnostic selection: tokens are chosen without conditioning on the {current} decoding step, retaining globally high-scoring items even when locally irrelevant. 
This neglect of step-wise relevance can degrade quality, especially as the generation focus drifts.

\paragraph{Design opportunities.}
These issues motivate an algorithm–system co-design method that (i) avoids separate score state, (ii) operates at page-aligned granularity, and (iii) performs context-aware, step-wise reconstruction, thereby preserving batching and translating sparsity into wall, clock speedups while maintaining quality.



\section{Methodology}
\label{sec:method}
In this section, we present {CHESS} (illustrated in Figure~\ref{fig:system_arch}), a context-aware, page-aligned KV selection framework. 
At each decoding step, CHESS uses a \emph{coarse-to-fine} hierarchy, {Grid} $\rightarrow$ {Chunk} $\rightarrow$ {Page}, to identify a budgeted set of relevant pages and \emph{reconstructs} the working KV set \emph{zero-copy}. 
Grid and Chunk are logical groupings for pruning; they share the same physical page storage.

\paragraph{High-level workflow.}
Selection proceeds top-down: CHESS first filters at the Grid level to shrink the search space, then refines at Chunk and Page levels. 
For decoding, it assembles the context by uniting the selected Pages with fixed \emph{attention sinks}~\cite{xiao2024efficientstreaminglanguagemodels} and the most recent tokens in the query window.
Details of the hierarchical summaries and the selection policy appear below.

\subsection{Hierarchical Semantic Representation}
\label{sec:hierarchy}

CHESS organizes the KV cache into a three-tier hierarchy: \textit{Pages}, \textit{Chunks}, and \textit{Grids}.
The fundamental unit is the \textit{Page} ($p_i$), consisting of $B$ tokens.
\textit{Page} is strategically aligned with the physical memory blocks used in PagedAttention~\cite{kwon2023efficientmemorymanagementlarge}, allowing the system to reference memory efficiently via page indices and utilize modern attention kernels.
Building on this, we define a \textit{Chunk} as a sequence of $N_{c}$ consecutive pages, and a \textit{Grid} groups $N_{g}$ chunks.

For semantic retrieval, we first derive a compact vector $\mathbf{v}_{p_i}$ for each page by aggregating its constituent Key states.
Specifically, we perform mean-pooling across the $B$ tokens and flatten the multi-layer, multi-head states into a single vector:
\begin{equation}
    \mathbf{v}_{p_i} = \operatorname{Flatten}\left( \left[ \frac{1}{B} \sum_{t \in p_i} \mathbf{K}^{(l,h)}_{t} \right]_{l, h} \right),
    \label{eq:page-vector}
\end{equation}
where $\mathbf{v}_{p_i}$ preserves dominant semantic features via high-dimensional orthogonality  (detailed proof in Appendix~\ref{app:theoretical_analysis}).

By flattening states across all layers, we capture the model's \textit{holistic latent state}, integrating both low-level syntactic features and high-level reasoning signals.
This comprehensive representation prevents the false positives often caused by analyzing only a single layer.

Finally, we derive representations for Chunks and Grids via hierarchical centroid averaging (see Figure~\ref{fig:system_arch}(a)):
\begin{equation}
    \mathbf{v}_{c_j} = \frac{1}{N_{c}} \sum_{p_i \in c_j} \mathbf{v}_{p_i}, \quad
    \mathbf{v}_{g_k} = \frac{1}{N_{g}} \sum_{c_j \in g_k} \mathbf{v}_{c_j}.
    \label{eq:hierarchy-vector}
\end{equation}
Although simple averaging may risk information dilution in lower dimensions, we rely on the \textit{hyper-dimensional sparsity} of LLM representations.
In such high-dimensional spaces, distinct semantic signals tend to remain orthogonal.
Consequently, the averaged vector $\mathbf{v}_{g_k}$ effectively preserves the dominant semantic information of the grid without catastrophic interference.

\subsection{Coarse-to-Fine Selection Mechanism}
\label{subsec:hierarchical-pruning}

To capture the evolving semantic information of the current generation, CHESS constructs a query anchor $\mathbf{v}_{anchor}$ by aggregating the \textit{Key states} of the recent local window $\mathcal{W}$:
\begin{equation}
    \mathbf{v}_{anchor} = \frac{1}{|\mathcal{W}|} \sum_{p_m \in \mathcal{W}} \mathbf{v}_{p_m}.
    \label{eq:query-vector}
\end{equation}
By utilizing Key states for both the query anchor and the historical context, we align the retrieval metric with the storage representation.
This ensures that we retrieve contexts that share the same latent feature distribution as the ongoing generation.

To supply the attention mechanism with the most relevant context, CHESS utilizes a metric termed \textit{Key-Key Semantic Affinity}. 
Leveraging the property that Key states inherently encode the semantic attributes of the input tokens, we compute the dot-product similarity between the current anchor and historical segments. 
A high affinity score indicates strong semantic alignment with the ongoing generation, thereby justifying the retention of these segments for the subsequent precise attention computation.
Formally, for any candidate unit $u$ (Grid, Chunk, or Page), we quantify its relevance to the current anchor $\mathbf{v}_{anchor}$ via:
\begin{equation}
    S(u) = \mathbf{v}_{anchor} \cdot \mathbf{v}_u.
    \label{eq:scoring}
\end{equation}
This formulation allows us to implement \textit{context-aware} selection using highly optimized matrix multiplication.
Guided by this metric, the selection policy executes in a top-down cascade governed by retention ratios $\{\rho_g, \rho_c, \rho_p\}$.
These hyperparameters are empirically calibrated to optimize the trade-off between generation quality and memory overhead.

{At the coarse level}, the system scans all grids in $\mathcal{G}$, retaining only the top $\rho_g$ fraction (denoted as $\mathcal{G}_{selected}$) to rapidly prune semantically irrelevant regions.
{Subsequently}, fine-grained verification is performed {conditionally}: chunks are evaluated \textit{if and only if} their parent grid resides in $\mathcal{G}_{selected}$. From this subset, the top $\rho_c$ fraction is preserved.
{At the finest granularity}, pages within these retained chunks are ranked, and the top $\rho_p$ fraction constitutes the semantic working set $\mathcal{P}_{selected}$.

To guarantee generation stability and prevent perplexity degradation, we augment this set by strictly preserving \textit{attention sinks}~\cite{xiao2024efficientstreaminglanguagemodels} and the most recent $\mathcal{W}$ pages.
The final decoding context is constructed as the union of $\mathcal{P}_{selected}$ and these safety pages.

\begin{algorithm}[!tb]
\caption{Batched Hierarchical Similarity Pruning}
\label{alg:batched_pruning}
\begin{algorithmic}[1]
\REQUIRE 
    Query Anchor $\mathbf{v}_{anchor} \in \mathbb{R}^{D}$; 
    Hierarchical Key Vectors $\{\mathbf{V}_g, \mathbf{V}_c, \mathbf{V}_p\}$; 
    Index Mappings $\mathcal{M}_{c \to g}, \mathcal{M}_{p \to c}$;
    Retention Ratios $\rho_g, \rho_c, \rho_p$.
\ENSURE Selected Page Indices $\mathcal{I}_{page}$

\STATE \textit{\footnotesize // Stage 1: Tensor Coalescing \& Global Scoring}
\STATE $\mathbf{V}_{all} \leftarrow \text{Concat}(\mathbf{V}_g, \mathbf{V}_c, \mathbf{V}_p)$ \label{alg:line:concat} \hfill \textit{\footnotesize // Stack hierarchies into unified tensor}
\STATE $\mathbf{S}_{all} \leftarrow \mathbf{v}_{anchor} \cdot \mathbf{V}_{all}^\top$ \label{alg:line:gemm} \hfill \textit{\footnotesize // Single GEMM Kernel Launch}
\STATE $\mathbf{S}_g, \mathbf{S}_c, \mathbf{S}_p \leftarrow \text{Split}(\mathbf{S}_{all})$ \hfill \textit{\footnotesize // Split scores back to levels}

\STATE \textit{\footnotesize // Stage 2: Hierarchical Masking (Vectorized)}
\STATE \textit{// Level 1: Grid Selection}
\STATE $\tau_g \leftarrow \text{Quantile}(\mathbf{S}_g, 1-\rho_g)$
\STATE $\mathbf{M}_g \leftarrow (\mathbf{S}_g \ge \tau_g)$ \hfill \textit{\footnotesize // Boolean Mask for Grids}

\STATE \textit{// Level 2: Chunk Selection (Conditional)}
\STATE $\mathbf{P}_{active}^{(c)} \leftarrow \text{Gather}(\mathbf{M}_g, \mathcal{M}_{c \to g})$ \label{alg:line:mask_start} \hfill \textit{\footnotesize // Propagate parent grid status}
\STATE $\mathbf{S}_c' \leftarrow \mathbf{S}_c \masked \mathbf{P}_{active}^{(c)}$ \hfill \textit{\footnotesize // Zero-out scores of inactive parents}
\STATE $\tau_c \leftarrow \text{Quantile}(\mathbf{S}_c', 1-\rho_c)$
\STATE $\mathbf{M}_c \leftarrow (\mathbf{S}_c \ge \tau_c) \land \mathbf{P}_{active}^{(c)}$ \label{alg:line:mask_end}

\STATE \textit{// Level 3: Page Selection (Conditional)}
\STATE $\mathbf{P}_{active}^{(p)} \leftarrow \text{Gather}(\mathbf{M}_c, \mathcal{M}_{p \to c})$ \hfill \textit{\footnotesize // Propagate parent chunk status}
\STATE $\mathbf{S}_p' \leftarrow \mathbf{S}_p \masked \mathbf{P}_{active}^{(p)}$
\STATE $\tau_p \leftarrow \text{Quantile}(\mathbf{S}_p', 1-\rho_p)$
\STATE $\mathbf{M}_p \leftarrow (\mathbf{S}_p \ge \tau_p) \land \mathbf{P}_{active}^{(p)}$

\RETURN $\text{ NonZero}(\mathbf{M}_p)$ \hfill \textit{\footnotesize // Return indices of active pages}
\end{algorithmic}
\end{algorithm}

\subsection{Adaptive Re-computation Strategy}
\label{sec:adaptive-strategy}

Relying solely on partial context selection involves inherent risks, as retrieving incorrect or irrelevant context can significantly degrade generation performance.
To mitigate this, CHESS incorporates a \textit{Quality-Aware Backtracking} mechanism that initiates context reconstruction strictly when generation quality deteriorates.
Specifically, CHESS monitors real-time generation dynamics via two complementary uncertainty metrics: average \textit{Entropy} ($\bar{H}_p$), which serves as a proxy for the model's lack of confidence, and \textit{Varentropy} ($\operatorname{Var}(H)_p$), which captures the temporal instability often characteristic of hallucination loops~\cite{kuhn2023semanticuncertaintylinguisticinvariances}.

This design ensures efficiency as reconstruction is bypassed as long as the generation remains stable.
To determine the triggering thresholds, we perform offline calibration on the golden dataset to analyze the joint distribution of Entropy and Varentropy.
We further demonstrate the superiority of this event-triggered approach over periodic context reconstruction.
For a detailed comparative analysis between our dynamic backtracking strategy and static reconstruction, please refer to Appendix~\ref{app:reconstruction_time}.



\section{System Implementation}
\label{sec:impl}

We implement CHESS atop \texttt{nanoVLLM}~\cite{nanovllm}.

\subsection{Hierarchical Structure Implementation}
\label{sec:hierarchy-impl}

\paragraph{Data structures for hierarchical organization.}
Our system manages the KV cache at the granularity of a \textit{page}, serving as the atomic unit for memory allocation.
\textit{Chunks} and \textit{Grids} are {logical views} as they restructure access patterns without physically duplicating the underlying tensor data.
This design maintains three distinct structural perspectives (Grid, Chunk, Page) over the same memory footprint, achieving a zero-copy implementation that incurs negligible memory overhead.

\subsection{High-Performance Pruning Implementation}
\label{sec:pruning_impl}

\paragraph{Batched similarity computation via tensor coalescing.}
To maximize GPU utilization and minimize kernel launch overhead, we vectorize the similarity computation across all hierarchy levels (Algorithm~\ref{alg:batched_pruning}).
Instead of sequentially iterating through Grids, Chunks, and Pages, we coalesce their semantic vectors into a unified tensor $\mathbf{V}_{all} \in \mathbb{R}^{N_{total} \times D}$ (Line 2), where $N_{total}$ sums the counts of all hierarchy nodes and $D$ denotes the feature dimension.
Utilization of the single query anchor $\mathbf{v}_{anchor}$ enables us to compute similarity scores for the entire hierarchy via a single {GEMM} operation: $\mathbf{S}_{all} = \mathbf{v}_{anchor} \cdot \mathbf{V}_{all}^\top$ (Line 3).
The subsequent filtering logic is implemented as a vectorized dependency check.
Specifically, a Chunk is selected only if its similarity score satisfies the threshold {and} its parent Grid is active, a condition enforced via hierarchical boolean masking (Lines 10 -- 13).
This design effectively replaces expensive control-flow divergence with efficient tensor operations.

\subsection{Integration with FlashInfer}
\label{sec:flashinfer-integration}

\paragraph{FlashInfer kernel.}
We leverage \texttt{FlashInfer}~\cite{ye2025flashinferefficientcustomizableattention} as our underlying attention engine to ensure high throughput while maintaining architectural flexibility.
A primary motivation for this choice is its support for {fine-grained page sizes} (e.g., 16 tokens).
In contrast, FlashAttention~\cite{dao2022flashattentionfastmemoryefficientexact} typically optimizes for coarser granularities (requiring block sizes to be multiples of 256), which introduces a {granularity mismatch} for semantic pruning.




\section{Experiments}
\label{sec:eval}

In this section, we evaluate {CHESS} along two dimensions: (i) \emph{generation quality} on long-context tasks and (ii) \emph{system efficiency} (throughput/latency) under varying sequence lengths.
Quality is measured on LongBenchV2~\cite{bai2025longbenchv2deeperunderstanding}; efficiency is measured on synthesized workloads across a spectrum of input lengths.


\subsection{Experimental Setup}
\label{sec:setup}

\begin{figure}[tbp]
    \centering
    \includegraphics[width=.9\linewidth]{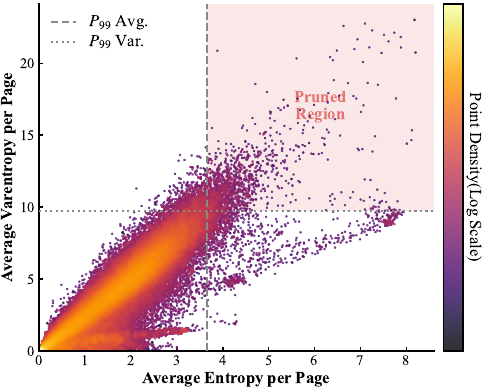}
    \caption{\textbf{Distribution of Average Entropy vs. Varentropy per KV Cache Page (Calibration Phase).} This plot illustrates the density of pages from the calibration dataset, where warmer colors indicate higher concentration. Dashed lines represent the selected 99th percentile thresholds. The shaded upper-right region highlights the \textit{pruned area}, corresponding to high-uncertainty outliers excluded by our method.}
    \label{fig:app-avg-vs-var}
    \vspace{-0.5cm}
\end{figure}

\begin{table*}[!htb]
\centering
\setlength{\tabcolsep}{5.5pt}
\caption{Main results on LongBench-v2. We report overall performance as well as breakdowns by difficulty (Easy, Hard) and context length (Short, Medium, Long). We compare CHESS against the best configurations of baselines: \texttt{H2O} (10\% heavy + 10\% local), \texttt{KeyDiff} (overall budget 8192 tokens), \texttt{SnapKV} (recent 512 tokens, total 4096 tokens), and \texttt{Quest} (budget 2048). For \texttt{CHESS}, we report three settings: Conservative (0.9, 0.9, 0.9), Moderate (0.8, 0.7, 0.7), and Aggressive (0.5, 0.2, 0.1).}
\begin{tabular}{l|c|c|cc|*{3}{w{c}{1.3cm}}}
\toprule
\multirow{2}{*}{\textbf{Method}} & \textbf{KV Cache} & \multirow{2}{*}{\textbf{Overall}} & \multicolumn{2}{c|}{\textbf{Difficulty}} & \multicolumn{3}{c}{\textbf{Length($<$32K;32K$\sim$128K;$>$128K)}} \\
 & \textbf{Budget} & & \textbf{Easy} & \textbf{Hard} & \textbf{Short} & \textbf{Medium} & \textbf{Long} \\ 
\midrule
\rowcolor{gray!10}\texttt{FullKV Inference} & 100\% & 30.2 & 33.9 & 28.0 & 38.3 & 24.2 & 28.7 \\
\midrule
\texttt{H2O (Best)} & \textbf{20\%} & \textbf{34.0} & \textbf{40.1} & 30.2 & 41.7 & \textbf{28.8} & 31.5 \\
\texttt{KeyDiff (Best)} & 8192 toks & 29.2 & 33.3 & 26.7 & 30.6 & 26.5 & 32.4 \\
\texttt{SnapKV (Best)} & 4096 toks & 30.2 & 34.9 & 27.3 & 34.4 & 24.2 & \textbf{35.2} \\
\texttt{Quest (Best)} & 2048 toks & 32.0 & 34.9 & 30.2 & 40.0 & 25.6 & 31.5 \\
\midrule
\texttt{CHESS (Conservative)} & 73\% & 30.4 & 34.9 & 27.7 & 36.1 & 26.0 & 29.6 \\
\texttt{CHESS (Moderate)} & 40\% & 32.2 & 35.4 & 30.2 & \textbf{41.7} & 24.7 & 31.5 \\
\rowcolor{blue!5} \textbf{\texttt{CHESS (Aggressive)}} & \textbf{1\%} & \underline{33.2} & \underline{38.0} & \textbf{30.2} & \underline{40.0} & \underline{27.4} & \underline{33.3} \\ 
\bottomrule
\end{tabular}
\label{tab:longbench-v2}
\end{table*}

\paragraph{Hardware and software environment.}
\label{par:environment}

All experiments were conducted on a single node with four H20 GPUs, using Python 3.12.3, PyTorch 2.5.1, and CUDA 12.4.

\paragraph{Baselines.}
\label{par:baselines}
We compare our method against \textbf{Full-KV}, the standard baseline that retains the complete KV cache during decoding to ensure lossless performance. 
To evaluate efficiency, we also include several state-of-the-art sparse attention mechanisms: \textbf{H2O}~\cite{zhang2023h2oheavyhitteroracleefficient} maintains a budget of ``heavy hitter'' tokens based on accumulated attention scores, while \textbf{KeyDiff}~\cite{keydiff} selects tokens by analyzing the distinctiveness of key distributions. 
Furthermore, \textbf{SnapKV}~\cite{li2024snapkvllmknowslooking} compresses context by identifying significant attention clusters to filter out redundant information. 
Finally, \textbf{Quest}~\cite{tang2024questqueryawaresparsityefficient} adopts a query-aware pruning strategy, dynamically estimating token importance to select a minimal subset for each decoding step.

To ensure a fair comparison, we integrated all baselines, except Quest, into CHESS evaluation stack. 
Quest relies on specialized kernels strictly optimized for single-batch inference, adapting them to our framework proved non-trivial. 
Consequently, we utilized its official implementation for benchmarking to ensure accurate evaluation.


\paragraph{Hyperparameters and configuration.}
\label{par:hyperparams}
Our system configuration relies on several key hyperparameters governing memory structure and sparsity levels. 
We standardize the \textbf{Page size} to 32 across all experiments. 
To control retention rates, we evaluate multiple configurations for \textbf{Grid}, \textbf{Chunk}, and \textbf{Page ratios}. 
Regarding uncertainty metrics, we perform offline calibration using the LongBenchV2 dataset~\cite{bai2025longbenchv2deeperunderstanding}. 
Specifically, we calculate the empirical distribution of entropy and varentropy, adopting the \textbf{99th percentile values} as the cutoff thresholds to prune high-uncertainty outliers. 
Figure~\ref{fig:app-avg-vs-var} illustrates the distribution and threshold.

\vspace{-0.2cm}
\subsection{Quality Evaluation}
\label{sec:quality}
Table~\ref{tab:longbench-v2} reports the results on LongBenchV2~\cite{bai2025longbenchv2deeperunderstanding}. 
Due to space constraints, only representative results are presented here; please refer to Appendix~\ref{app:comprehensive_table} for the full evaluation.
With only {1\%} of the KV cache, {CHESS (Aggressive)} achieves an overall score of {33.2}, outperforming {FullKV} (30.2) and remaining close to {H2O} (34.0), which uses {20$\times$} more KV budget.
This indicates CHESS preserves task-relevant evidence while effectively filtering redundant or distracting context that can harm long-context reasoning.

\begin{figure}[!tb]
    \centering
    \vspace{-0.7cm}
    \includegraphics[width=.9\linewidth]{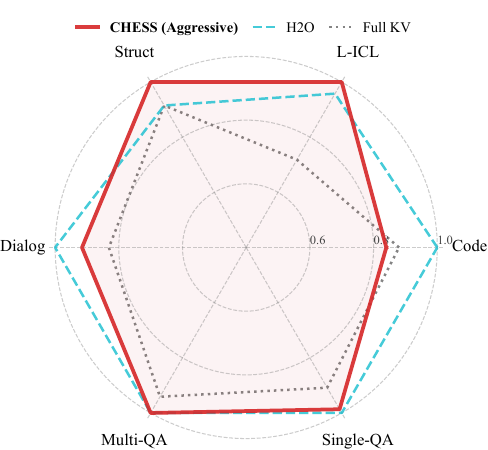}
    \caption{\textbf{Normalized accuracy on long-context tasks.} Axes represent relative performance scaled to the maximum score per domain. Abbreviations: \textbf{L-ICL}: Long In-context Learning; \textbf{Struct}: Structured Data; \textbf{Dialog}: Dialogue History; \textbf{Multi/Single-QA}: Multi/Single-Doc QA.}
    \label{fig:radar}
    \vspace{-0.3cm}
\end{figure}

Figure~\ref{fig:radar} breaks down performance by domain.
CHESS consistently matches or exceeds {FullKV} across all categories, with particularly strong gains on {Single-QA}, {Multi-QA}, {L-ICL}, and {Structured} tasks.
These tasks benefit from {coherent, contiguous context}, where CHESS’s context-aware, page-level selection retains semantically aligned regions rather than isolated tokens.
In contrast, accumulation-based methods tend to preserve globally frequent but locally irrelevant tokens, which can dilute effective context.

{H2O} achieves its best results on {Code}, where relevant statements are often non-contiguous and sparsely distributed.
In such cases, H2O’s attention-accumulation mechanism is better to capture sparse dependencies.
Despite this, CHESS remains competitive on Code while delivering substantial gains on the majority of long-context reasoning tasks, demonstrating a more favorable quality–efficiency trade-off.

Overall, these results show that CHESS attains near state-of-the-art quality under extreme KV budgets by reconstructing {semantically coherent context} at each decoding step, rather than relying on global token importance.

\begin{figure*}[!htbp]
    \centering
    \includegraphics[width=.95\textwidth]{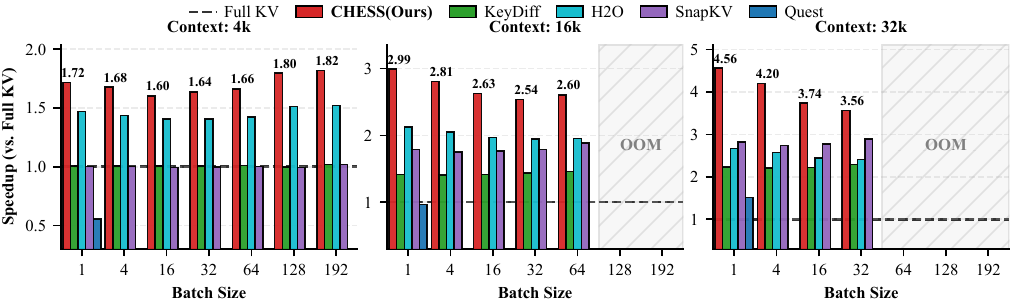}
    \caption{\textbf{End-to-end Throughput Speedup Comparison.} We report the throughput ratios of \textbf{CHESS} and alternative sparse attention methods relative to the \textbf{Full-KV} baseline (normalized to $1.0\times$, indicated by the dashed line). The evaluation spans varying input context lengths ($4\text{k}$--$32\text{k}$) and batch sizes ($1$--$192$). CHESS consistently achieves the highest throughput, peaking at a $\mathbf{4.56\times}$ speedup in the $32\text{k}$ context setting. Gray shaded regions indicate Out-of-Memory (OOM) scenarios. Note that \textbf{Quest} is evaluated only at batch size 1 due to implementation constraints limiting it to single-batch inference.}
    \label{fig:throughput}
\end{figure*}

\subsection{System Performance}
\label{sec:performance}

We evaluate system efficiency using the {CHESS (Aggressive)} configuration across varying context lengths and batch sizes.
Our results show that CHESS consistently outperforms prior sparse-KV methods under realistic long-context workloads, with performance gains that \emph{increase} as the workload becomes more demanding.

\paragraph{Throughput analysis.}
\label{par:throughput}
Figure~\ref{fig:throughput} displays the end-to-end throughput normalized to Full-KV inference. 
Corresponding results for other GPU architectures are provided in Appendix~\ref{app:throughput}.
CHESS achieves the highest throughput across all evaluated settings, peaking at a {4.56$\times$} speedup under a 32k context.
Notably, the performance gap between CHESS and existing methods widens with increasing batch size and context length.
While prior sparse approaches exhibit diminishing returns as batch size grows, CHESS continues to scale effectively.
This indicates that CHESS alleviates fundamental system bottlenecks, rather than providing workload-specific optimizations.

\begin{figure}[!htb]
    \centering
    \vspace{-0.5cm}
    \includegraphics[width=\linewidth]{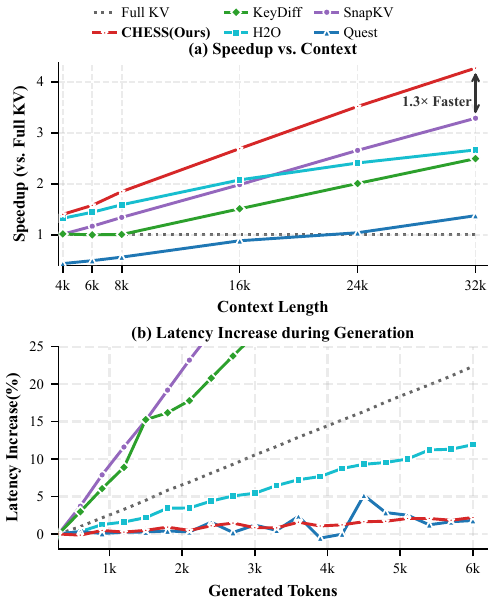}
    \caption{
        (a) \textbf{TPOT Speedup Scalability across Context Lengths ($4\text{k}$--$32\text{k}$).} 
        CHESS demonstrates superior scalability, with the performance gap widening significantly as sequence length increases. 
        At the $32\text{k}$ context, CHESS achieves a peak speedup of $\mathbf{4.56\times}$ over Full KV, outperforming the strongest baseline (SnapKV) by $\mathbf{1.3\times}$. 
        (b) \textbf{Latency Stability during Long-Sequence Generation.} 
        The plot tracks the latency increase over $6\text{k}$ generated tokens (following a $32\text{k}$ input). 
        While baselines exhibit linear growth (Full KV) or severe instability (SnapKV/KeyDiff), CHESS maintains a consistent, flat latency profile with negligible overhead.
    }
    \vspace{-0.1cm}
    \label{fig:latency-scalability}
\end{figure}

\paragraph{Long input/output tasks (scalability and stability ).}
\label{par:latency-scalability}
Figure~\ref{fig:latency-scalability}(a) reports the scalability of Time Per Output Token (TPOT) speedup as context length increases.
CHESS consistently widens the performance gap over all baselines, with the advantage becoming more pronounced at longer contexts.
This trend reflects the {dual bottleneck} of long-context decoding—memory bandwidth pressure from KV transfers and growing attention computation.
By retaining only a sparse, context-relevant KV cache, CHESS alleviates both factors simultaneously, achieving up to $\mathbf{4.3\times}$ speedup over Full KV and outperforming the strongest baseline (SnapKV) by $\mathbf{1.3\times}$ at $32\text{k}$ context.

Figure~\ref{fig:latency-scalability}(b) evaluates latency stability during long-sequence generation. 
As generation progresses, the KV cache grows, leading to steadily increasing computation and memory traffic per token.
Accordingly, {FullKV} and {SnapKV} exhibit a clear linear increase in per-token latency as the effective cache expands.
In contrast, {CHESS} maintains a nearly constant decoding latency throughout generation.
By retaining the most contextually relevant KV entries, CHESS effectively decouples per-token latency from sequence length.

\begin{figure}[!tbp]
    \centering
    \includegraphics[width=.9\linewidth]{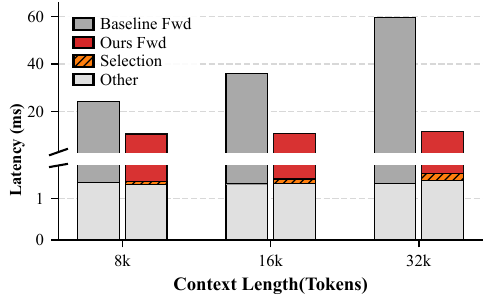}
    \caption{\textbf{Latency breakdown per decoding step.} Comparison between Full KV and CHESS (Aggressive) with batch size 1. The "Other" category includes scheduling, sampling, and data preparation. The overhead from the CHESS's selection mechanism is negligible relative to the total decoding latency.}
    \label{fig:latency_breakdown_appendix}
\end{figure}

\paragraph{Selection overhead.}
Figure~\ref{fig:latency_breakdown_appendix} presents a per-step latency breakdown.
The additional overhead introduced by CHESS’s selection mechanism is visually negligible.
Quantitatively, selection accounts for only {0.72\%}, {0.98\%}, and {1.49\%} of total latency at 8k, 16k, and 32k contexts, respectively.
This minimal overhead is achieved through two factors:
(1) {amortized execution}, where selection is triggered only when necessary, and
(2) {kernel-efficient implementation} using optimized GEMM operations.
As a result, CHESS delivers substantial system-level speedups without introducing new performance bottlenecks.
\section{Related Work}
\label{sec:related}


We review prior work along two dimensions: (i) retention strategy and (ii) selection metric and processing granularity
A comprehensive taxonomy is detailed in Appendix~\ref{app:taxonomy}.

\subsection{Retention Strategy}
\label{subsec:static-dynamic}

\textbf{Static eviction.} 
“Prune-once” methods build the retained context at prefill and keep it fixed during decoding.
{SnapKV}~\cite{li2024snapkvllmknowslooking} identifies important patterns during the prefilling stage and keeps them for later generations.
{KeyDiff}~\cite{keydiff} utilizes key matrix similarity to filter tokens. 
While these static methods significantly reduce memory footprint, their prompts are fixed regarding the generation phase. 
Once a token is evicted, it cannot be recovered, leading to information loss when the generation requires revisiting historically pruned content.

\textbf{Dynamic reconstruction.} 
To mitigate information loss, some propose dynamically constructing the context for each generation step. 
{H2O}~\cite{zhang2023h2oheavyhitteroracleefficient} maintains a set of ``Heavy Hitter'' tokens based on accumulated attention scores, dynamically updating the cache. 
{RocketKV}~\cite{rocketKV} further optimizes this by using approximated attention scores. 
{InfLLM}~\cite{infLLM} offloads KV pairs to CPU memory and retrieves blocks when required.
Dynamic schemes adapt to changing relevance, though they introduce selection state and runtime overheads.


\subsection{Selection Metric and Processing Granularity}
\label{subsec:metric-granularity}

\textbf{Selection metric.} 
Most existing methods (e.g., H2O, SnapKV, InfLLM) utilize attention score as the selection metric. 
This introduces significant system overhead, as maintaining accumulated attention scores requires additional HBM traffic and frequent updates, which can bottleneck the inference engine. 
In contrast, key-based approaches (\eg KeyDiff) utilize key semantics. 
Since key matrices are already resident in HBM for attention computation, using them as a semantic probe avoids the overhead of storing and updating auxiliary score matrices.

\textbf{Processing granularity.} 
Methods operating at the token level face system-level inefficiencies. 
At the \emph{token} level, pruning yields non-contiguous memory access patterns that disrupt memory coalescing and make it harder to integrate with kernel optimizations designed for PagedAttention~\cite{kwon2023efficientmemorymanagementlarge}. 
By contrast, \emph{block/page}–level methods (\eg ChunkKV~\cite{chunkkv}, Quest~\cite{tang2024questqueryawaresparsityefficient}) operate on contiguous pages, naturally aligning with paging in modern inference engines and enabling zero-copy selection with better hardware utilization.

In summary, prior approaches leave three gaps: (i) \emph{context-agnostic} selection that ignores step-wise relevance, (ii) irrecoverable \emph{static} eviction or \emph{dynamic} token-level schemes that rely on auxiliary score state, and (iii) a system misalignment where irregular access patterns and score maintenance hinder batching and kernel efficiency. 

\section{Conclusion}

We propose CHESS, an algorithm–system co-design for KV-cache management in long-context LLMs. 
CHESS departs from context-agnostic pruning by introducing a context-aware, hierarchical selection strategy that reconstructs a coherent working set at each decoding step, while its page-aligned, zero-copy execution translates algorithmic sparsity into practical wall-clock gains.
Extensive experiments on LongBenchV2 and large-scale synthetic workloads show that CHESS preserves generation quality at the level of Full-KV inference while using only {1\%} of the KV cache.
At the system level, CHESS consistently improves throughput and latency, achieving up to {4.56$\times$} speedup over Full-KV and outperforming strong state-of-the-art baselines in long-context regimes.
In the future, we plan to integrate CHESS with speculative decoding and RAG-augmented pipelines, and explore richer uncertainty signals for selection.


\section*{Impact Statement}

This paper presents work whose goal is to advance the field of Machine Learning through improved efficiency in long-context language model inference. There are many potential societal consequences of our work, none which we feel must be specifically highlighted here.



\bibliography{example_paper}
\bibliographystyle{icml2026}

\newpage
\appendix
\onecolumn

\section{Taxonomy of existing methods}
\label{app:taxonomy}

Table~\ref{tab:related-work} summarizes representative KV-cache management methods along five orthogonal dimensions: selection metric, processing granularity, context awareness, dynamic reconstruction, and self-correction capability.
This taxonomy highlights several systematic gaps in prior work.

First, most existing approaches rely on \emph{context-agnostic} signals, such as attention scores or positional heuristics, to identify important tokens.
While effective for reducing cache size, these metrics do not condition on the semantic intent of the \emph{current} decoding step, which limits their ability to adapt to shifting generation focus.

Second, a large fraction of prior methods operate at the \emph{token level}.
Although fine-grained, token-level selection often conflicts with block-based memory management in modern inference engines, complicating efficient execution.
Block-level methods partially address this issue, but still largely depend on attention-based metrics and remain context-agnostic.

Third, only a subset of methods supports \emph{dynamic reconstruction} of the KV cache, and none explicitly provide a mechanism to \emph{recover} from erroneous pruning decisions once relevant context has been removed.
As a result, quality degradation is often irreversible under aggressive sparsity settings.

In contrast, \textbf{CHESS} uniquely combines (i) key-based semantic signals, (ii) block-level selection aligned with paged KV storage, (iii) explicit context awareness, (iv) dynamic reconstruction, and (v) a self-correcting mechanism.
This combination enables CHESS to achieve aggressive cache reduction while maintaining generation quality and system efficiency, distinguishing it from prior approaches.

\begin{table*}[htbp]
    \centering
    \caption{Comparison with related works. \textbf{Metric}: The criterion used to identify important tokens; \textbf{Granularity}: The processing unit (Token vs. Block); \textbf{Ctx. Aware}: Whether selected tokens are semantically relevant to the current generation context; \textbf{Dyn. Recon.}: Whether the method dynamically reconstructs the KV cache context during generation; \textbf{Self-Correct}: Whether the method employs a mechanism to recover from retrieval errors.}
    \label{tab:related-work}
    \resizebox{\linewidth}{!}{
    \begin{tabular}{lccccc}
    \toprule
    \textbf{Method} & \textbf{Selection Metric} & \textbf{Granularity} & \textbf{Ctx. Aware} & \textbf{Dyn. Recon.} & \textbf{Self-Correct} \\
    \midrule
    StreamingLLM~\cite{xiao2024efficientstreaminglanguagemodels} & Position & Token & -- & -- & -- \\
    H2O~\cite{zhang2023h2oheavyhitteroracleefficient} & Attn Score & Token & -- & \checkmark & -- \\
    RocketKV~\cite{rocketKV} & Attn Score & Token & -- & \checkmark & -- \\
    KeyDiff~\cite{keydiff} & Key Matrix & Token & -- & -- & -- \\
    SnapKV~\cite{li2024snapkvllmknowslooking} & Attn Score & Token & -- & -- & -- \\
    ChunkKV~\cite{chunkkv} & Attn Score & Block & -- & -- & -- \\
    InfLLM~\cite{infLLM} & Attn Score & Block & -- & \checkmark & -- \\
    Quest~\cite{tang2024questqueryawaresparsityefficient} & Attn Score & Block & -- & \checkmark & -- \\
    \rowcolor{gray!15} \textbf{CHESS (Ours)} & \textbf{Key Matrix} & \textbf{Block} & \textbf{\checkmark} & \textbf{\checkmark} & \textbf{\checkmark} \\
    \bottomrule
    \end{tabular}
    }
\end{table*}

\section{Theoretical Analysis of CHESS}
\label{app:theoretical_analysis}

\subsection{Semantic Preservation of Hierarchical Pooling}
The core of CHESS relies on representing a Page ($B$ tokens) by its mean-pooled key vector $v_p$. A potential concern is whether this pooling operation dilutes critical semantic signals.

\paragraph{High-Dimensional Orthogonality} 
In the high-dimensional latent space of LLMs (where $D$ is typically 4096 or larger), random vectors tend to be nearly orthogonal. 
Formally, for any two unrelated key vectors $K_i$ and $K_j$, their inner product $K_i \cdot K_j \approx 0$. 
When we compute the mean-pooled vector $v_p = \frac{1}{B} \sum_{i=1}^{B} K_i$, the unrelated "noise" tokens cancel each other out due to this orthogonality, while the dominant semantic signal (the "Heavy Hitter") persists.

\paragraph{Selection Consistency} 
We define the Key-Key Semantic Affinity as $S(u) = v_{anchor} \cdot v_u$. Leveraging the Concentration Inequality, it can be shown that if a specific token $K_i$ within a page has a high affinity with the query, the aggregate page vector $v_p$ will also maintain a high score with high probability:
\begin{equation}
P\left( |v_{anchor} \cdot v_p - v_{anchor} \cdot K_i| > \epsilon \right) \leq 2\exp\left( - \frac{C \cdot B \cdot \epsilon^2}{\sigma^2} \right).
\end{equation}
This ensures that our hierarchical selection (Grid $\to$ Chunk $\to$ Page) maintains high recall, as important pages are unlikely to be pruned at coarser levels.

\subsection{Complexity and System Efficiency}
CHESS transforms the linear scanning cost of long-context inference into a budgeted hierarchical search.

\paragraph{Computational Complexity} 
Traditional Full-KV inference requires $O(L)$ memory reads and attention computations. 
In contrast, CHESS's hierarchical filtering reduces the active KV set size. 
By coalescing semantic vectors into a unified tensor $V_{all}$, we compute similarity for the entire hierarchy via a single GEMM kernel launch. 
This reduces the effective per-step complexity to $O(\rho \cdot L)$, where $\rho$ is the retention ratio (e.g., 1\%).

\paragraph{Zero-Copy Logic} 
Unlike token-level pruning that requires expensive physical data movement to maintain memory continuity, CHESS operates at a page-aligned granularity. 
By manipulating logical page indices within the PagedAttention framework, we achieve "selection-on-the-fly" without any data movement overhead, directly translating theoretical sparsity into wall-clock speedups.

\section{Choice of reconstructing the context}
\label{app:reconstruction_time}

\begin{figure}[h]
    \centering
    \includegraphics[width=.8\linewidth]{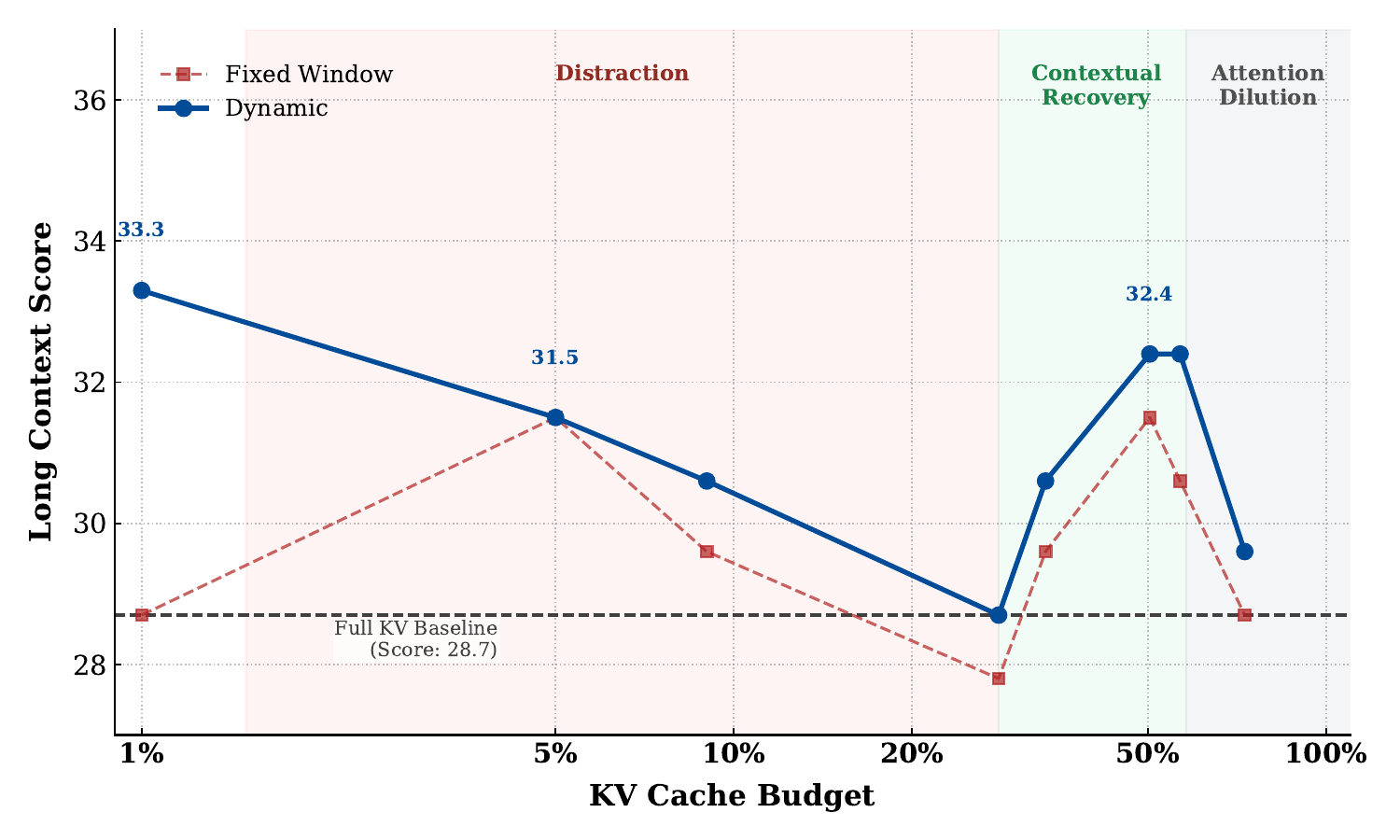} 
    \caption{Comparison between Fixed Window and Dynamic Construction on LongBenchV2. The trend highlights the phases of Distraction, Recovery, and Dilution.}
    \label{fig:dynamic_vs_fixed}
\end{figure}

We evaluate the efficiency of our dynamic mechanism by comparing it against a static triggering configuration. 
To ensure a rigorous evaluation, we configured the baseline to reconstruct the context every 6 pages—a frequency significantly higher than our dynamic approach (averaging $\sim$10 pages). 
Despite the baseline benefiting from more frequent updates, the results in Figure~\ref{fig:dynamic_vs_fixed} demonstrate that our dynamic construction consistently yields superior performance across various budget ratios.

Crucially, the performance trajectory reveals a non-monotonic trend characterized by three distinct phases: \textit{Distraction}, \textit{Contextual Recovery}, and \textit{Attention Dilution}. 
At a strict 1\% budget, the method achieves peak efficiency by isolating only the most precise, high-value information, consistent with recent observations that optimized sparse caches can outperform full-context baselines \cite{evolkv}. 
However, as the budget increases moderately, the system enters the \textit{Distraction} phase: the cache begins to admit lower-rank tokens but lacks the capacity to retain their surrounding semantic context. 
Consequently, these isolated tokens act as noise that distracts the attention mechanism, leading to a temporary drop in quality \cite{easily_distract}. 
Subsequently, when the budget reaches a critical threshold, the \textit{Contextual Recovery} phase begins; the cache gains enough capacity to restore the necessary background for these tokens, thereby recovering model reasoning. 
Finally, at very high budgets, we observe \textit{Attention Dilution}, where attention scores become overly diffuse, causing performance to converge toward the Full KV baseline, mirroring limitations observed in extended context processing \cite{litm}.

\section{Throughput results on other GPUs}
\label{app:throughput}
In this section, we provide additional throughput evaluations conducted on a node equipped with NVIDIA A800 GPUs. Note that \textbf{Quest} is omitted from this comparison as its current implementation is restricted to a batch size of 1 and does not support multi-batch inference scenarios.

\begin{figure*}[!htbp]
    \centering
    \includegraphics[width=\textwidth]{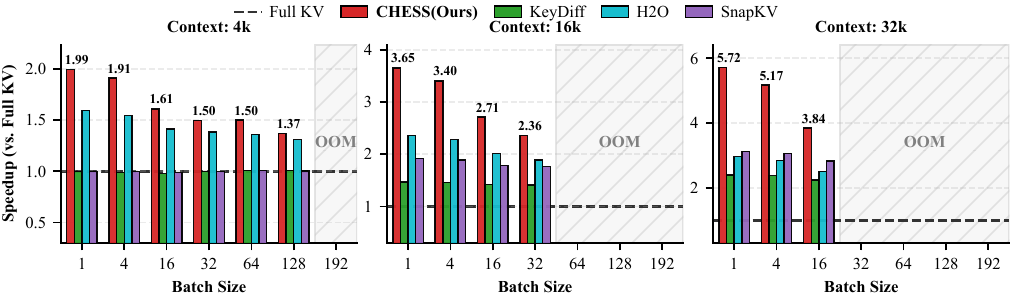}
    \caption{End-to-end throughput speedup on a multi-GPU system (4$\times$ NVIDIA A800). We report the normalized throughput of \textbf{CHESS} and other representative sparse attention baselines. All results are relative to the \textbf{Full-KV} baseline (normalized to $1.0\times$, represented by the horizontal dashed line). Higher values indicate better efficiency in processing long sequences.}
    \label{fig:throughput_comparison_a800}
\end{figure*}

Figure~\ref{fig:throughput_comparison_a800} reports end-to-end throughput speedups on a multi-GPU system equipped with NVIDIA A800 GPUs, complementing the main results obtained on H20/A100-class hardware.
Across all evaluated context lengths (4k, 16k, and 32k) and batch sizes, \textbf{CHESS} consistently achieves the highest throughput among all compared methods.

Several observations are worth noting.
First, the relative performance trends closely mirror those observed on other GPU platforms.
In particular, the throughput advantage of CHESS \emph{widens} as context length increases, reaching up to $5.72\times$ speedup over Full-KV at 32k context.
This confirms that the benefits of context-aware, block-aligned KV selection are not hardware-specific, but persist across different GPU architectures.

Second, CHESS maintains robust scalability with respect to batch size.
While baselines such as H2O and SnapKV experience diminishing returns or early out-of-memory (OOM) failures as batch size grows, CHESS sustains high throughput even under large-batch, long-context workloads.
This behavior highlights the practical advantage of CHESS's page-level, zero-copy execution, which avoids the memory fragmentation and synchronization overheads that limit token-level approaches.

Finally, although all baselines are evaluated relative to the same Full-KV reference, CHESS consistently delivers higher gains even under this normalized setting.
This suggests that the observed speedups stem from fundamental improvements in KV-cache efficiency rather than favorable hardware configurations or benchmark artifacts.

Overall, these results demonstrate that CHESS generalizes well across GPU platforms and system configurations, reinforcing its applicability as a robust KV-cache management strategy for long-context LLM inference.

\section{LongBenchV2 table}
\label{app:comprehensive_table}
While the main text (cf. Table~\ref{tab:longbench-v2}) provides a representative subset of results, this section presents a more comprehensive evaluation of CHESS and other baselines across a wider spectrum of dynamic ratio configurations on LongBench-v2. 
By sweeping through various budget settings, we aim to demonstrate the robustness of CHESS under diverse resource constraints.
The full results are summarized in Table~\ref{tab:longbench-v2-comprehensive}. 

CHESS maintains strong performance across a wide range of cache budgets, from moderately compressed settings (73\%–40\%) down to extremely sparse regimes (9\% and 1\%).
Notably, even at aggressive compression levels (e.g., 9\% and 1\%), CHESS remains competitive with or surpasses baselines that retain substantially larger KV caches.
This indicates that CHESS is not sensitive to a narrow operating point, but instead offers a broad and stable quality–efficiency trade-off.

In contrast, compared methods like KeyDiff and SnapKV exhibit sharp performance degradation as the budget decreases, particularly on hard and long-context subsets.
This behavior reflects the brittleness of context-agnostic or token-isolated selection.
By reconstructing semantically coherent context blocks, CHESS degrades gracefully under extreme sparsity and avoids catastrophic quality collapse.

At the most constrained setting (1\% KV budget), CHESS achieves the highest overall score among all sparse methods, and even surpasses Full-KV inference.
This suggests that CHESS not only preserves essential information but can also filter redundant or distracting context, improving effective reasoning in long-context scenarios.

Across Easy/Hard and Short/Medium/Long splits, CHESS demonstrates balanced gains without overfitting to a particular regime.
In particular, its advantage is pronounced on long-context inputs, aligning with the core motivation of context-aware reconstruction.

\begin{table*}[!htb]
\centering
\setlength{\tabcolsep}{5.5pt}
\caption{Evaluation of CHESS and baselines under diverse dynamic ratio configurations.}
\begin{tabular}{l|c|c|cc|*{3}{w{c}{1.3cm}}}
\toprule
\multirow{2}{*}{\textbf{Method}} & \textbf{KV Cache} & \multirow{2}{*}{\textbf{Overall}} & \multicolumn{2}{c|}{\textbf{Difficulty}} & \multicolumn{3}{c}{\textbf{Length($<$32K;32K$\sim$128K;$>$128K)}} \\
 & \textbf{Budget} & & \textbf{Easy} & \textbf{Hard} & \textbf{Short} & \textbf{Medium} & \textbf{Long} \\ 
\midrule
\rowcolor{gray!10}\texttt{FullKV Inference} & 100\% & 30.2 & 33.9 & 28.0 & 38.3 & 24.2 & 28.7 \\
\midrule
\texttt{H2O} & 20\% & 34.0 & 40.1 & 30.2 & 41.7 & 28.8 & 31.5 \\
\texttt{H2O} & 40\% & 33.0 & 38.5 & 29.6 & 42.2 & 27.4 & 28.7 \\
\texttt{H2O} & 60\% & 31.4 & 34.9 & 29.3 & 41.7 & 23.7 & 29.6 \\
\texttt{KeyDiff} & 1024 toks & 22.9 & 25.5 & 21.2 & 28.9 & 16.7 & 25.0 \\
\texttt{KeyDiff} & 2048 toks & 24.9 & 29.7 & 21.9 & 30.0 & 19.5 & 26.9 \\
\texttt{KeyDiff} & 4096 toks & 27.4 & 32.8 & 24.1 & 32.8 & 23.3 & 26.9 \\
\texttt{KeyDiff} & 8192 toks & 29.2 & 33.3 & 26.7 & 30.6 & 26.5 & 32.4 \\
\texttt{SnapKV} & 512 toks & 16.5 & 17.7 & 15.8 & 19.4 & 13.0 & 18.5 \\
\texttt{SnapKV} & 1024 toks & 24.5 & 29.7 & 21.2 & 30.0 & 20.9 & 22.2 \\
\texttt{SnapKV} & 4096 toks & 30.2 & 34.9 & 27.3 & 34.4 & 24.2 & 35.2 \\
\texttt{Quest} & 1024 toks & 31.4 & 34.9 & 29.3 & 36.7 & 26.0 & 33.3 \\
\texttt{Quest} & 2048 toks & 32.0 & 34.9 & 30.2 & 40.0 & 25.6 & 31.5 \\
\texttt{Quest} & 4096 toks & 30.1 & 41.3 & 23.1 & 43.4 & 20.6 & 30.4 \\
\midrule
\texttt{CHESS} & 73\% & 30.4 & 34.9 & 27.7 & 36.1 & 26.0 & 29.6 \\
\texttt{CHESS} & 57\% & 31.0 & 34.4 & 28.9 & 40.6 & 24.7 & 27.8 \\
\texttt{CHESS} & 50\% & 31.2 & 34.9 & 28.9 & 38.3 & 25.6 & 30.6 \\
\texttt{CHESS} & 40\% & 32.2 & 35.4 & 30.2 & 41.7 & 24.7 & 31.5 \\
\texttt{CHESS} & 28\% & 32.2 & 37.0 & 29.3 & 41.7 & 25.1 & 30.6 \\ 
\texttt{CHESS} & 21\% & 31.6 & 36.5 & 28.6 & 40.0 & 24.7 & 31.5\\ 
\texttt{CHESS} & 9\% & 31.8 & 35.4 & 29.6 & 38.9 & 26.5 & 30.6 \\ 
\texttt{CHESS} & 1\% & 33.2 & 38.0 & 30.2 & 40.0 & 27.4 & 33.3 \\ 
\bottomrule
\end{tabular}
\label{tab:longbench-v2-comprehensive}
\end{table*}

\end{document}